\documentclass[10pt,twocolumn,letterpaper]{article}

\usepackage{cvpr}
\usepackage{times}
\usepackage{epsfig}
\usepackage{graphicx}
\usepackage{amsmath}
\usepackage{amssymb}

\usepackage[breaklinks=true,bookmarks=false]{hyperref}

\cvprfinalcopy % *** Uncomment this line for the final submission

\usepackage{algorithm}
\usepackage{algorithmic}
\usepackage{booktabs}
\usepackage{enumerate}
\usepackage{multirow}
\usepackage{subfigure}
\def\cvprPaperID{2295} % *** Enter the CVPR Paper ID here

\ifcvprfinal\fi
\begin{document}

%%%%%%%%% TITLE
\title{Data Augmentation Revisited:\\Rethinking the Distribution Gap between Clean and Augmented Data}

\author{Zhuoxun He \textsuperscript{\rm 1}\quad  Lingxi Xie \textsuperscript{\rm 2}\quad  Xin Chen \textsuperscript{\rm 3}\quad  Ya Zhang \textsuperscript{\rm 1} \quad Yanfeng Wang  \textsuperscript{\rm 1}\quad  Qi Tian \textsuperscript{\rm 2}\\ 
\textsuperscript{\rm 1} Shanghai Jiao Tong University \quad \textsuperscript{\rm 2}Huawei Noah’s Ark Labc \quad \textsuperscript{\rm 3}Tongji University\\
\{zhuoxun, ya\_zhang, wangyanfeng\}@sjtu.edu.cn \quad 198808xc@gmail.com \\
chenxin061@hotmail.com \quad tian.qi1@huawei.com
}

% \and

\maketitle
%\thispagestyle{empty}

%%%%%%%%% ABSTRACT
\begin{abstract}
Data augmentation has been widely applied as an effective methodology to improve generalization in particular when training deep neural networks. Recently, researchers proposed a few intensive data augmentation techniques, which indeed improved accuracy, yet we notice that these methods augment data have also caused a considerable gap between clean and augmented data. In this paper, we revisit this problem from an analytical perspective, for which we estimate the upper-bound of expected risk using two terms, namely, empirical risk and generalization error, respectively. We develop an understanding of data augmentation as regularization, which highlights the major features. As a result, data augmentation significantly reduces the generalization error, but meanwhile leads to a slightly higher empirical risk. On the assumption that data augmentation helps models converge to a better region, the model can benefit from a lower empirical risk achieved by a simple method, \textit{i.e.}, \textbf{using less-augmented data to refine the model trained on fully-augmented data}. Our approach achieves consistent accuracy gain on a few standard image classification benchmarks, and the gain transfers to object detection. 

\end{abstract}

%%%%%%%%% BODY TEXT
\section{Introduction}

% Recent years have witnessed a rapid development of deep learning approaches. 
With the availability of powerful computational resources nowadays, it is possible to train very deep neural networks that have been verified to achieve good performance in a wide range of computer vision tasks including image classification~\cite{krizhevsky2012imagenet,Szegedy_2015_CVPR,he2016deep,he2019bag}, object detection~\cite{ren2015faster,lin2017feature}, semantic segmentation~\cite{badrinarayanan2017segnet,deeplabv3plus2018}, \textit{etc}. On the other hand, complicated models with tens of millions of parameters are often difficult to train with limited training data, and many techniques~\cite{srivastava2014dropout,ioffe2015batch} have been proposed to stable training and improve generalization.

\begin{figure}[t]
\centering
\includegraphics[width=8.25cm]{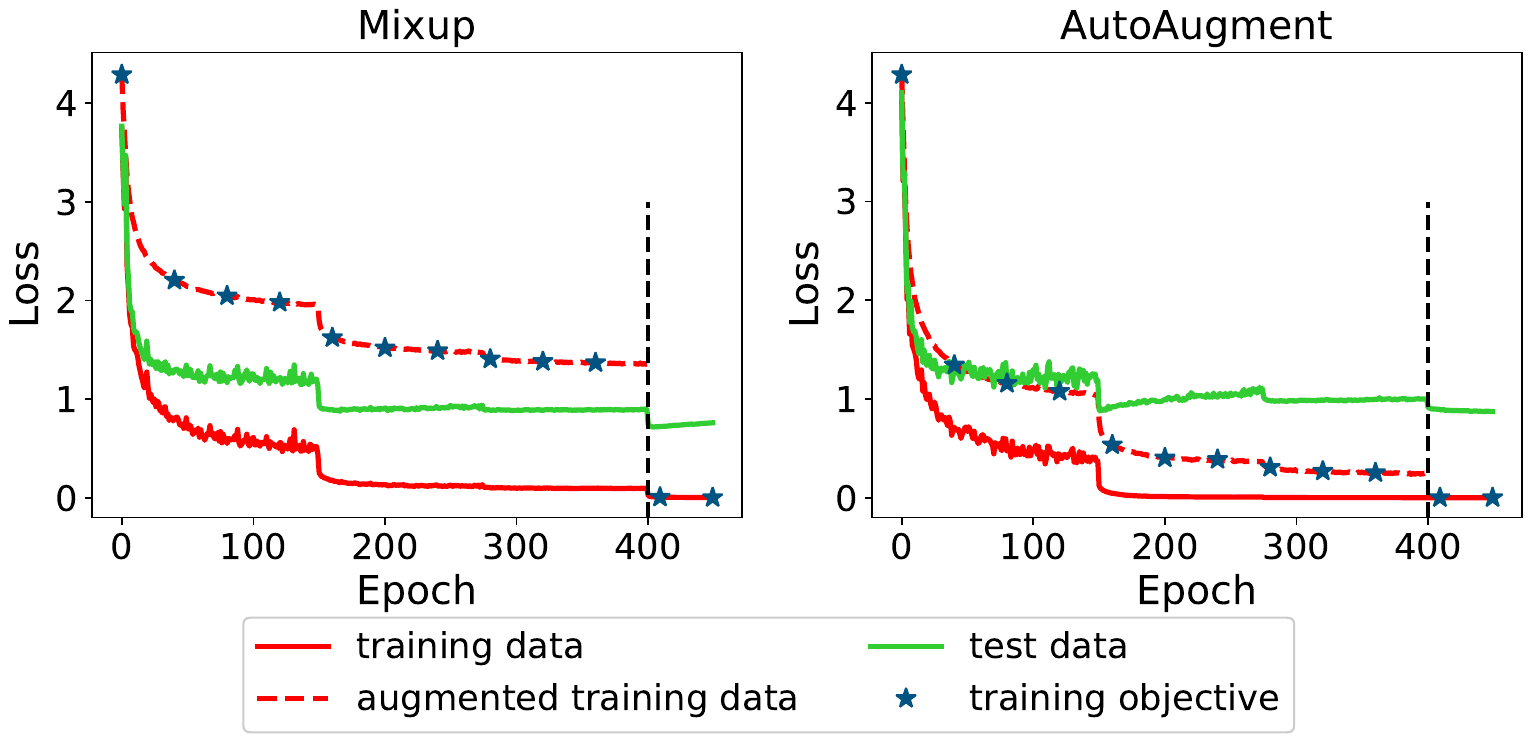}
    % {\includegraphics[width=4cm,height=3.2cm]{mixup.pdf}}
    % {\includegraphics[width=4cm,height=3.2cm]{autoaug.pdf}}
\caption{Cross-entropy loss curves in training \textit{PreActResNet-18} on CIFAR100, with two popular data augmentation methods, namely, Mixup and AutoAugment. Mind the gap between clean and augmented \textit{training} data, while the model is trained on augmented data during whole conventional training process. As indicated by $\star$, our approach is to train the model on augmented data in the first 400 epochs, and on original data in the final 50 epochs, which helps to achieve lower testing losses, \textit{i.e.}, higher accuracy.}
\label{fig:illustration}
\vspace{-0.2cm}
\end{figure}

As a common training trick, data augmentation is designed to increase the diversity of training data without actually collecting new data. Essentially, it is initially driven by the fact that slight modification on an image will not change the high-level semantics of an image. Standard data augmentation methods used in deep network training involves scale and aspect ratio distortions, random crops, and horizontal flips. However, several sophisticated methods changing the original image a lot, such as Mixup~\cite{zhang2017mixup}, Cutout~\cite{devries2017improved}, and Cutmix~\cite{yun2019cutmix}, were proposed to improve the generalization of models. Recently, researchers also used AutoML to learn an optimal set of parameters for augmentation~\cite{Cubuk_2019_CVPR,lim2019fast}, and achieved state-of-the-art performance in a few image classification benchmarks.

With these complicated data augmentation methods being proposed, quite high training loss, even higher than the testing loss, can be got at the end of the training stage. The question of why models with such high training loss can generalize well remains mostly uncovered. This paper delves deep into this issue by reformulating the upper-bound of expected risk for models trained with data augmentation, which implies it is that the empirical risk on original data, but not augmented data, and the generalization error impact the performance of models. Then the effectiveness of intensive data augmentation is equivalent to how to guarantee the convergence of the empirical risks when a large distribution gap between clean and augmented data. By partitioning features into major and minor features, we develop an understanding of intensive data augmentation as regularization on the minor features to guarantee the convergence of the empirical risk on clean data. Even so, the model trained on augmented data has a slightly higher empirical risk in general while generalizes better. 

Motivated by this observation, we propose an effective training strategy to reduce the empirical risk and improve performance further on the assumption that models trained with data augmentation arrive in a better region. Compared to conventional training strategy in which augmented data are used during the whole process, our approach is to switch off these intensive data augmentation methods at the last few training epochs. This is to confirm that there is no intensive distribution gap between clean and augmented data at the end of the training stage, which we believe is a safe opportunity of transferring the trained model to the testing scenario. As shown in Figure~\ref{fig:illustration}, by plotting the curves during the training process, we observe that refining the models without intensive data augmentation decreases the empirical risk further, and meanwhile, a lower testing loss is achieved.

We evaluate our training strategy in a few popular image classification benchmarks including CIFAR10, CIFAR100, Mini-ImageNet, and ImageNet. In all these datasets, the refined model achieves consistent, sometimes significant, accuracy gain, with computational costs in training and testing remain unchanged. Besides, we fine-tune the pre-trained models on ImageNet on PascalVOC following the standard strategy. The pre-trained models with our approach show a significant improvement in the detection task, while it was believed these intensive data augmentation methods were inappropriate for detection. Beyond image recognition experiments, our research sheds light on a new optimization method, which involves using augmented data followed by clean data to optimize global and local properties, and is worth being further studied in the future.

%-------------------------------------------------------------------------
\section{Related Work}
The statistical learning theory \cite{vapnik1998statistic} suggested that the generalization of the learning model was characterized by the capacity of the model and the amount of training data.
Data augmentation methods have been commonly adopted in deep learning by creating artificial data to improve model robustness, especially in image classification \cite{krizhevsky2012imagenet,Szegedy_2015_CVPR}. Basic image augmentation methods mainly include elastic distortions, scale, translation, and rotation \cite{ciregan2012multi,sato2015apac}. For natural image datasets, random cropping and horizontal flipping are common. Besides, scaling hue, saturation, and brightness also improve performance \cite{he2019bag}. These methods were usually designed by retaining semantic information unchanged while generating various images. 

Recently, some methods to synthesize training data were proposed. Mixup~\cite{zhang2017mixup,tokozume2018between} combined two samples linearly in pixel level, where the target of the synthetic image was the linear combination of one-hot label encodings. There are a few variants of Mixup~\cite{guo2018mixup,verma2018manifold}, as well as a recent effort named Cutmix~\cite{yun2019cutmix} which combined Mixup and Cutout~\cite{devries2017improved} by cutting and pasting patches. Compared to moderate data augmentation, images generated by Mixup and its variants are much more different from original data and unreal to human perception.
% even though those methods try to decrease the distribution gap between clean and synthetic data. 

For different datasets, the best augmentation strategies can be different. Learning more sensible data augmentations for specific datasets has been explored~\cite{lemley2017smart,ratner2017learning,tran2017bayesian}. AutoAugment~\cite{Cubuk_2019_CVPR} used a search algorithm based on reinforcement learning to find the best data augmentation strategies from a discrete search space that is composed of augmentation operations. Population Based Augmentation~\cite{jaderberg2017population,ho2019pba} proposed to search for augmentation policy schedules instead of fixed augmentation policies. Furthermore, Fast AutoAugment~\cite{lim2019fast} largely accelerated AutoAugment by avoiding training on each policy. More recently, augmentation in the deep feature space by intra-class covariance matrices was efficiently implemented with a new loss function~\cite{wang2019implicit}.

Compared to numerous works in the empirical aspect, relatively few theoretical works explained the effectiveness of data augmentation, especially in deep learning. Bishop~\cite{bishop1995training} showed that training with noise had the effect of regularization in expectation. Rajput~\etal analyzed the performance of data augmentation through the lens of margin~\cite{rajput2019does}. Dao~\etal~\cite{dao2019kernel} connected data augmentation with kernels, and show that data augmentation as feature averaging and variance regularization. In this paper, we take a different path to analyze the impacts of data augmentation from the upper bound of expected risk.

The generalization ability of DNN is also greatly affected by optimization methods~\cite{hardt2016train,zhang2017understand,keskar2017improving}. Even with some regularization methods~\cite{srivastava2014dropout,ioffe2015batch}, the objective function can not converge to the global minimum. The DNN tends to learn low-level features firstly which can be easier to learn for machines such as size, color, and texture~\cite{gatys2015texture,GATYS2017texture,geirhos2018imagenettrained}. Brendel and Bethge~\cite{brendel2018approximating} found that the decision-making behavior of current neural network architectures was mainly based on relatively weak and local features, and high-level features, \textit{e.g.}, shape, that can better improve model robustness were not sufficiently learned. 
% This paper associates data augmentation with optimization methods and proposes a simple yet effective approach to improve the performance of models. 
Recent works~\cite{achille2017critical,golatkar2019time} found that regularization does not need to exist in the whole training process, which is compatible with our work. 

%-------------------------------------------------------------------------

\section{Data Augmentation Revisited}\label{sec_data}

In this section, we revisit data augmentation by identifying the loss terms that compose the upper-bound of the expected risk. Then, we provide an explanation on how data augmentation works as regularization, followed by a simple and practical approach to improve the performance of models trained by data augmentation.
\subsection{Statistical Learning with Data Augmentation}
Let $\mathcal{X}$ and $\mathcal{Y}$ be the data and label spaces, respectively. Each sample is denoted by $(x,y) \sim \mathcal{P}$, where $\mathcal{P}$ is the joint distribution of data and label. Consider a regular statistical learning process. The goal of learning is to find a function $f:\mathcal{X}\mapsto\mathcal{Y}$ which minimizes the expected value of a pre-defined loss term, $\ell(f(x),y)$, over the distribution of $\mathcal{P}$. This is known as the \textit{expected risk}:
$$
R(f|\mathcal{P})=\int \ell(f(x), y) \mathrm{d}P(x, y).
$$
However, the data distribution is unknown in practical situations. Therefore, a common solution is the \textit{empirical risk minimization} (ERM) principle~\cite{vapnik1998statistic}, which optimizes an \textit{empirical risk} in a training dataset $\left\{(x_n,y_n)\right\}_{n=1}^N$ that mimics the data distribution:
$$
\hat{R}(f|\mathcal{P})=\frac{1}{N} \sum_{n=1}^{N} \ell\left(f\left(x_{n}\right), y_{n}\right).
$$

The accuracy of estimating the expected risk goes up with $N$, the amount of training data. In practice, especially when there is limited data, increasing $N$ with data augmentation is a popular and effective solution. It defines a function $g\in\mathcal{G}:\mathcal{X}\mapsto\mathcal{X}$, which generates `new' data $\tilde{x}_n$ with a combination of operations on $x_n$ -- since these operations do not change the semantics, $\tilde{x}_n$ naturally shares the same label with $x_n$, \textit{i.e.}, $\tilde{y}_i=y_n$ Note that data augmentation actually changes the distribution of $\mathcal{P}$ and we denote the new distribution by $\mathcal{P}_\mathrm{aug}$, which is to say, the goal has been changed from minimizing $\hat{R}(f|\mathcal{P}))$ to minimizing $\hat{R}(f|\mathcal{P}_\mathrm{aug})$:
$$
\hat{R}(f|\mathcal{P}_\mathrm{aug})=\frac{1}{N} \sum_{n=1}^{N} \ell\left(f\left(\tilde{x}_{n}\right), \tilde{y}_{n})\right).
$$

The strategy of data augmentation can be conservative, in which only a small number of `safe' operations such as horizontal flip and cropping are considered, or aggressive, in which `dangerous' or a series of operations can be used to cause significant changes on image appearance. Here we briefly introduce several aggressive data augmentation methods, all of which were proposed recently and verified effectiveness in image classification tasks.

For $(x,y)$ and $(x',y') \sim \mathcal{P}$, the generated data $(\tilde{x}, \tilde{y})$ by Mixup~\cite{zhang2017mixup} is obtained as follows:
\begin{equation}
\tilde{x}=\lambda \cdot x+(1-\lambda) \cdot x',\ \ \tilde{y}=\lambda \cdot y+(1-\lambda)\cdot y',
\end{equation}
where $\lambda \sim \mathrm{Beta}(\gamma,\gamma)$ and $\gamma$ is the combination ratio (a hyperparameter). Manifold Mixup~\cite{verma2018manifold} randomly performs the linear combination at an eligible layer that can be input layer and some hidden layer.

%\textcolor{red}{More aggressively}, 
Cutmix~\cite{yun2019cutmix} combines CutOut~\cite{devries2017improved} and Mixup, providing a patch-wise, weighted overlay by:
\begin{equation}
\tilde{x}=\mathbf{M} \odot x+(\mathbf{1}-\mathbf{M}) \odot x',\ \ \tilde{y}=\lambda \cdot y+(1-\lambda)\cdot y',
\end{equation}
where $\mathbf{M}$ is a binary mask indicating the positions of drop-out and fill-in, $\odot$ denotes element-wise multiplication and $\lambda \sim \mathrm{Beta}(1,1)$ is the combination ratio.

% \subsubsection{AutoAugment} 
Instead of manually designing data augmentation tricks, AutoAugment~\cite{Cubuk_2019_CVPR} applied an automated way of learning parameters for augmentation. A large space with different kinds of operations was pre-defined, and the policy of augmentation was optimized with reinforcement learning. This is to say, the function $g$ applied to each sample for augmentation can be even more complicated compared to those in conventional approaches.

\subsection{Rethinking the Mechanism of Augmentation}

According to VC theory~\cite{vapnik1998statistic}, the consistency and generalization of ERM principle have been justified in theoretical aspect. Consider a binary classifier $f \in \mathcal{F}$, which has finite VC-Dimension $|\mathcal{F}|_{\mathrm{VC}}$. With probability $1-\delta$, a upper bound of the \textit{expected} risk is formulated by 
\begin{equation}
    R(f) \leqslant \hat{R}(f)+O\left(\left(\frac{|\mathcal{F}|_{\mathrm{VC}}-\log \delta}{N}\right)^{\alpha}\right),
    \label{euq_1}
\end{equation}
where $\frac{1}{2} \leqslant \alpha \leqslant 1$. In the simply (separable) case, $\alpha=1$, and in the complex (non-separable) case, $\alpha=\frac{1}{2}$.

Based on this theory, data augmentation creates sensible data to increase the training data size. Assume there is a finite number of augmented training data. For the model trained over the augmented data distribution $\mathcal{P}_\mathrm{aug}$, we have
\begin{equation}
\begin{aligned}
    R(f_{\mathrm{aug}}|\mathcal{P}_\mathrm{aug}) \leqslant & \hat{R}(f_{\mathrm{aug}}|\mathcal{P}_\mathrm{aug}) + O\left(\left(\frac{|\mathcal{F}|_{\mathrm{VC}}-\log \delta}{M\times N}\right)^{\alpha}\right),
\end{aligned}
\label{euq_3}
\end{equation}
where $M$ is a finite constant.  

Although the generalization error in Equation~(\ref{euq_3}) is smaller than that in Equation~(\ref{euq_1}), there is a difference between other risk terms. Note $\mathcal{P} \subseteq \mathcal{P}_{\mathrm{aug}}$, and $\mathcal{P}_{\mathrm{aug}}$ can be more difficult to learned. Suppose that 
$$
R(f_{\mathrm{aug}}|\mathcal{P}_{\mathrm{aug}}) - R(f_{\mathrm{aug}}|\mathcal{P}) = \varepsilon_1 \geqslant 0,
$$
$$
\hat{R}(f_{\mathrm{aug}}|\mathcal{P}_{\mathrm{aug}}) - \hat{R}(f_{\mathrm{aug}}|\mathcal{P})= \varepsilon_2 \geqslant 0.
$$
Thus, the inequality is reformulated by
\begin{equation}
    R(f_{\mathrm{aug}}|\mathcal{P}) \leqslant \hat{R}(f_{\mathrm{aug}}|\mathcal{P}) + O\left(\left(\frac{|\mathcal{F}|_{\mathrm{VC}}-\log \delta}{M\times N}\right)^{\alpha}\right)+ \varepsilon,
\label{euq_4}
\end{equation}
where $\varepsilon = \varepsilon_2 - \varepsilon_1$. Since both $\varepsilon_1$ and $\varepsilon_2$ are caused by the distribution gap between $\mathcal{P}$ and $\mathcal{P}_\mathrm{aug}$, it is reasonable to assume $\varepsilon$ is sufficiently small. 

As a result, Equation (\ref{euq_4}) highlights that the benefits of learning with data augmentation mainly arise due to two factors:
\vspace{-0.05cm}
\begin{enumerate}[1)]
    \item \textit{the empirical risk $\hat{R}(f_{\mathrm{aug}}|\mathcal{P})$ being small,}
    \vspace{-0.25cm}
    \item \textit{the amount of augmented data being large,}
    % \vspace{-0.3cm}
    % \item \textit{the approximation error being small.}
\end{enumerate}
\vspace{-0.05cm}
The conclusions provide a deep insight, \textit{i.e.}, it is the value of $\hat{R}(f_{\mathrm{aug}}|\mathcal{P})$ but not $\hat{R}(f_{\mathrm{aug}}|\mathcal{P}_{\mathrm{aug}})$ impacts the effectiveness of data augmentation. 

Since the model is trained on $\hat{R}(f_{\mathrm{aug}}|\mathcal{P}_\mathrm{aug})$, factor 1) requires the consistency between $\hat{R}(f_{\mathrm{aug}}|P)$ and $\hat{R}(f_{\mathrm{aug}}|\mathcal{P}_\mathrm{aug})$. If no distribution gap exists, \textit{i.e.}, $\mathcal{P}=\mathcal{P}_\mathrm{aug}$, the consistency is guaranteed. Unfortunately, there is some kind of trade-off between the amount of augmented data and the distribution gap. Recent augmentation methods that generate many augmented images by changing images much in appearance can lead to an intensive distribution gap, such as Mixup, AutoAugment and so on. Intuitively, such an intensive distribution gap would greatly impact the convergence of $\hat{R}(f_{\mathrm{aug}}|\mathcal{P})$. However, previous empirical results demonstrate their effectiveness and verify the convergence of $\hat{R}(f_{\mathrm{aug}}|\mathcal{P})$. 

\subsection{Convergence of the Empirical Risk}\label{sec_converge}
The effectiveness of \textit{intensive} data augmentation can be described as the following question:

\vspace{0.15cm}
\textit{How can one guarantee the consistency between $\hat{R}(f_{\mathrm{aug}}|\mathcal{P})$ and $\hat{R}(f_{\mathrm{aug}}|\mathcal{P}_\mathrm{aug})$, when a large distribution gap between clean and augmented data exists?}
\vspace{0.15cm}

Let $\mathcal{H} \subseteq \mathbb{R}^D$ and $\mathbf{h}(x) = (h_1(x),h_2(x),\ldots,h_D(x))^{\top}$ denote latent space (feature space) and the feature vector of $x$ respectively. Suppose a perfect classifier is given by the true conditional distribution is $\mathcal{Q}(y|\mathbf{h})$. We partition the features into \textbf{major features} and \textbf{minor features} by information gain. For major features, the possibility density $q(y|h_d)$ concentrates on some point mass. For minor features, the possibility density $q(y|h_d)$ is relatively uniform.

To simplify the question, we explore it in conjunction with a linear softmax classifier  $\mathbf{W}=[\mathbf{w}_1, \mathbf{w}_2, ..., \mathbf{w}_C] \in \mathbb{R}^{D\times C}$, where $C$ is the number of categories. The predicted label is given by $\hat{y}=[p_1,p_2,...,p_C]$:
\begin{equation}
p_i = \frac{\exp (\mathbf{w}_i^{\top}\mathbf{h})}{\sum_{j=1}^{C}{\exp (\mathbf{w}_j^{\top}\mathbf{h})}} = \frac{1}{\sum_{j=1}^{C}{\exp ((\mathbf{w}_j-\mathbf{w}_i)^{\top}\mathbf{h})}}
\end{equation}

In the objective function on original data $\hat{R} = \frac{1}{N} \sum_{n=1}^{N} \ell\left(\mathbf{W}^{\top}\mathbf{h}\left(x_{n}\right), y_{n}\right)$, the related terms with some feature $h_d$ are $(w_{i,d}-w_{j,d})h_d$ for $1 \leqslant i,j \leqslant C$. When $h_d$ is a minor feature, the variation of $h_d$ should not change the results much, which requires the weights $|w_{i,d}-w_{j,d}|\rightarrow 0$. Further, this can be reformulated as $w_{i,d} \approx w_{j,d}$ for $1 \leqslant i,j \leqslant C$.

For intensive data augmentation $g\sim\mathcal{G}$ which brings a large distribution gap between clean and augmented data, we have $||\mathbf{h}(g(x))-\mathbf{h}(x)||_2 > \epsilon_0$, where $\epsilon_0 > 0$ is a relative large value.
Then, objective function with data augmentation is given as follow:
$$
\hat{R}_\mathrm{aug} = \frac{1}{N} \sum_{n=1}^{N} \mathbb{E}_{g\sim\mathcal{G}}\left[\ell\left(\mathbf{W}^{\top}\mathbf{h}\left(g(x_{n})\right), y_{n}\right)\right].
$$
As the analysis in~\cite{dao2019kernel}, we expand each term of the objective function with data augmentation using Taylor expansion:
\begin{equation}
\begin{aligned}
    \mathbb{E}_{g\sim \mathcal{G}}[\ell(\mathbf{W}^{\top}\mathbf{h}&\left(g(x)\right), y)]=\\ &\ell\left(\mathbf{W}^{\top}\bar{\mathbf{h}}, y\right) + 
    \frac{1}{2} \mathbb{E}_{g\sim \mathcal{G}}[\Delta^{\top}\mathbf{H}(\tau,y)\Delta]
\end{aligned}
\label{euq_taylor}
\end{equation}
where $\bar{\mathbf{h}} = \mathbb{E}_{g\sim \mathcal{G}}\left[\mathbf{h}(g(x))\right]$, $\Delta=\mathbf{W}^{\top}\left(\bar{\mathbf{h}}- \mathbf{h}(g(x))\right)$ and $\mathbf{H}$ is the Hessian matrix. Dao \etal~\cite{dao2019kernel} proposed data augmentation as feature averaging and variance regularization. Here we further discuss the effects of intensive data augmentation to emphasise the regularization on the corresponding weights of minor features.

For cross-entropy loss with softmax, the $\mathbf{H}$ is positive semi-definite and independent of $y$. Then, the second-order term in Equation (\ref{euq_taylor}) requires that $w_{i,d} \rightarrow 0$ for all $i$, if the variance of $h_d(g(x))$ is large. Since the intensive data augmentation must cause large variances of some features and such regularization on $w_{i,k}$ is not appropriate for major features, a reasonable solution is given:
\begin{enumerate}[1)]
    \item \textit{for major features,} $|h_d(g(x))-h_d(x)| < \zeta_1$,
    \vspace{-0.25cm}
    \item \textit{for minor features,} $|h_{d}(g(x))-h_{d}(x)| > \epsilon_1$,
\end{enumerate}
\vspace{-0.25cm}
\textit{where $\zeta_1 > 0$ is a small value, and $\epsilon_1 \geqslant \zeta_1$.}
\vspace{0.3cm}

These two inequalities highlight that the major features that are important to classify should be preserved as much and the minor features can be changed a lot after data augmentation. Comparing that $\hat{R}$ restricts $w_{i,d} \approx w_{j,d}$ for $1 \leqslant i,j \leqslant C$, $\hat{R}_{\mathrm{aug}}$ directly restricts $w_{i,d}\rightarrow 0$ for minor features $h_d$. While maintaining the optimized objective consistent, the intensive data augmentation also regularizes the corresponding weights of minor features. 

This is consistent with the empirical results \cite{Cubuk_2019_CVPR}, whose augmentation policies are selected by $\hat{R}$ on a validation set. For numeral recognition, the transformation invert is successful to be used, even though the numeral specific color is changed to that not involved in the original dataset. On the other hand, the transformation horizontal flipping used commonly in natural images is never used in numeral recognition. It is consistent with prior knowledge that the relative color of the numeral and its background and the direction of the numeral are major features, but the specific color of numeral is a minor feature. 

\subsection{Refined Data Augmentation}
As discussed in the last subsection, intensive data augmentation methods highlight the major features, while losing some minor features. In such way, data augmentation keeps the consistency of $\hat{R}(f|\mathcal{P})$ and $\hat{R}(f|\mathcal{P}_{\mathrm{aug}})$ and regularize the corresponding weights of minor features. From this perspective, data augmentation as a regularization scheme imposing some constraints on the function space $\mathcal{F}$ by prior knowledge, which forces the model to focus on the major features.

\begin{figure}[t]
\centering
{\includegraphics[width=4cm,height=3.3cm]{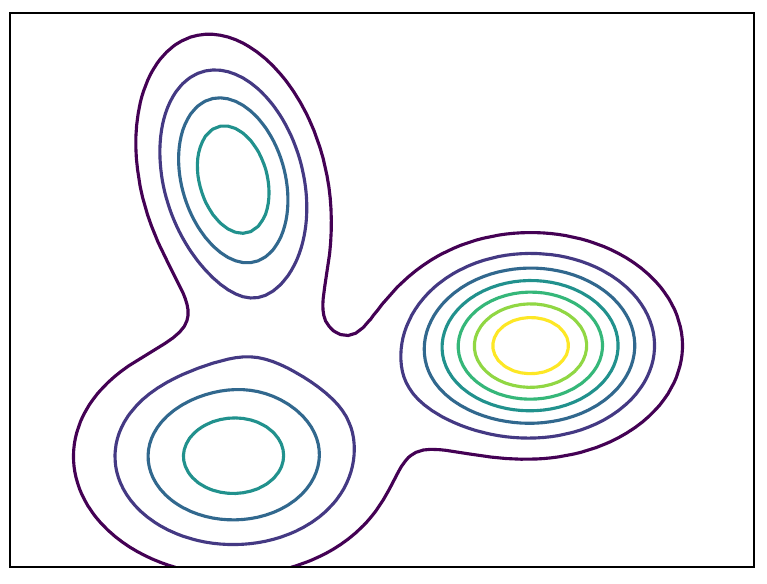}}
{\includegraphics[width=4cm,height=3.3cm]{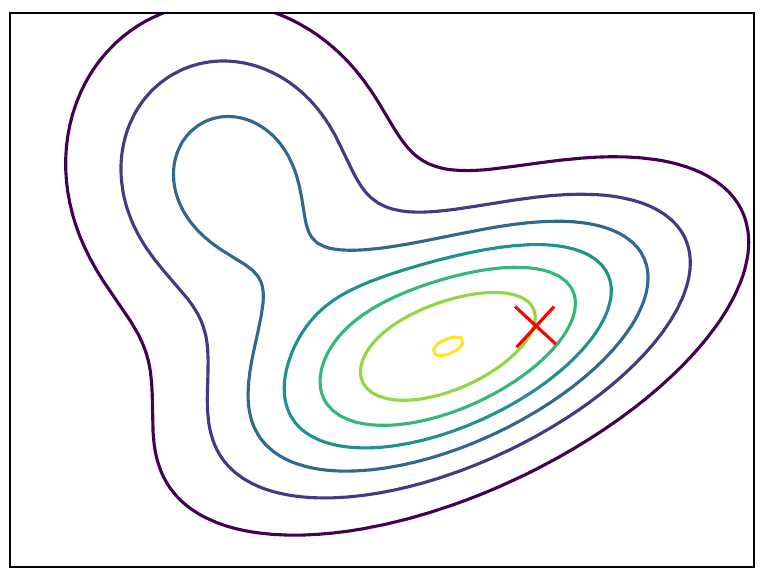}}
\caption{The contour maps of the empirical risks on the function space $\mathcal{F}$. 
\textbf{Left}: $\hat{R}(f|\mathcal{P})$. \textbf{Right}:$\hat{R}(f|\mathcal{P}_\mathrm{aug})$, where the ``$\times$'' denotes the global minimum of $\hat{R}(f|\mathcal{P})$.}
\label{fig_1}
\end{figure}

Figure~\ref{fig_1} depicts the effect of data augmentation on changing the empirical risk $\hat{R}(f)$ on the function space $\mathcal{F}$. Data augmentation helps the neural network learn major features, which reduces the number of local minimums and keeps the direction of convergence relatively consistent. Compared with $\hat{R}(f|\mathcal{P}_{\mathrm{aug}})$, $\hat{R}(f|\mathcal{P})$ can easily converge to a local minimum that is far away from the global minimum. 

To go a step further, since the analysis based on Equation (\ref{euq_taylor}) has introduced some approximation, the optimal function $f^{\dag}$ of  $\hat{R}(f|\mathcal{P}_{\mathrm{aug}})$ is not guaranteed to be a minimum of $\hat{R}(f|\mathcal{P})$. Associated the empirical results, $\hat{R}(f^*|\mathcal{P}) \leqslant \hat{R}(f^{\dag}|\mathcal{P})$ in general, where $f^*$ is optimized on $\hat{R}(f|\mathcal{P})$, while $f^{\dag}$ generalizes better. Please refer to Table~\ref{tab_cifar} and Table~\ref{tab_loss} for experimental verifications. Therefore, it is reasonable to believe that $f^{\dag}$ converges to a better region close to the global minimum.
% Moreover, the features in practice can have more precise weights for classifying, and those relatively minor features are indeed informative for classifying.

Motivated by this, we propose a method called \textbf{refined data augmentation}, \textit{i.e.}, refining the models without intensive data augmentation at the end of the training stage. On one hand, the distribution gap between clean and augmented data obstructs $\hat{R}(f|\mathcal{P})$ to converge further, and a small empirical risk can benefit the performance of models. On the other hand, the relatively minor features ignored by intensive data augmentation are also informative for classifying.

\section{Experiments}

%We evaluate our approach, refined data augmentation, on top of several popular augmentation methods, including Mixup~\cite{zhang2017mixup}, Manifold Mixup~\cite{verma2018manifold}, Cutmix~\cite{yun2019cutmix}, and AutoAugment~\cite{Cubuk_2019_CVPR}. The evaluations are based on several commonly used image classification benchmarks, and validated in the object detection task on PascalVOC~\cite{everingham2010pascal}.

\subsection{Results on CIFAR10 and CIFAR100}

\noindent$\bullet$\quad\textbf{Dataset and Settings}

The CIFAR10 both consist of 60,000 $32\times32$ color images in 10 classes, where 5,000 training images per class as well as 1,000 testing images per class. The CIFAR100 contains 500 training images and 100 testing images per class in a total of 100 classes. On CIFAR10 and CIFAR100, we train both two variants of residual networks~\cite{he2016deep}, PreActResNet-18~\cite{he2016identity} and WideResNet-28-10~\cite{BMVC2016_87}, and  a stronger backbones: Shake-Shake~\cite{gastaldi2017shake}. We partition the training procedure into two stages: training with intensive data augmentation and refinement. 

For the stage with intensive data augmentation, we train PreActResNet-18 and WideResNet-28-10 on a single GPU using PyTorch for 400 epochs on training set with a minibatch size of 128. For PreActResNet-18, the learning rate starts at 0.1 and is divided by 10 after 150, 275 epochs, and weight decay is set to be $10^{-4}$. For WideResNet-28-10, the learning rate starts at 0.1 and is divided by 5 after 120, 240, 320 epochs except for using a Cosine learning rate decay~\cite{loshchilov2016sgdr} for AutoAugment, and weight decay is set to be $5\times10^{-4}$. Following the settings in Zhang \etal~\cite{zhang2017mixup} and Cubuk \etal~\cite{Cubuk_2019_CVPR}, we set dropout rate to be 0.3 for WideResNet-28-10 with AutoAugment, and 0 in other experiments. For Shake-Shake model, we train the model on 2 GPUs for 1800 epochs with a mini-batch size of 128. The learning rate starts at 0.01 with Cosine decay, and weight decay is set to be $10^{-3}$.

All intensive data augmentation methods are incorporated with standard data augmentation: random crops, horizontal flips with a probability of 50\%. For the coefficient $\lambda \sim \mathrm{Beta}(\gamma, \gamma)$ in Mixup and Manifold Mixup, $\gamma =1$. Following the paper~\cite{yun2019cutmix}, Cutmix is implemented with a probability of 50\% during training. For AutoAugment, we first apply the standard data augmentation methods, then apply the AutoAugment policy, then apply Cutout with $16\times16$ pixels~\cite{devries2017improved} following Cubuk \etal~\cite{Cubuk_2019_CVPR}. Note that we directly use the AutoAugment policies reported in~\cite{Cubuk_2019_CVPR}.

% \subsubsection{Refinement}
We refine the models without these intensive data augmentation methods. Since the standard data augmentation methods bring a small distribution gap between clean and augmented data, we preserve the standard data augmentation when refining, which will be discussed in detail later. For PreActResNet-18 and WideResNet-28-10, the models are refined for 50 epochs, and for Shake-Shake, the models are refined for 200 epochs. During refinement, the learning rate keeps a small value. For the models trained with the step-wise learning rate decay, the learning rate is set to be the same as that in the final epoch of the last stage, and for the models trained with the Cosine learning rate decay, the learning is adjusted to a reasonably small value.

% \begin{table*}[]
% \centering
% \begin{tabular}{lccccccc}
% \toprule
% Networks        & N & M & Batch size & GPUs & Weight decay & Initila lr & lr schedule \\
% \midrule
% resnet18        & 400                   & 50                    & 128                            & 1                        & 1e-4                         & 0.1                            & 0.1 after 150, 275 epochs       \\
% wideresnet28-10 & 400                   & 50                    & 128                            & 1                        & 5e-4                         & 0.1                            & 0.2 after 120, 240, 320 epochs  \\
% Shake-shake     & 1800                  & 200                   & 128                            & 2                        & 1e-3                         & 0.01                           & consine decay  \\
% \bottomrule
% \end{tabular}
% \end{table*}

\begin{table}[!tb]
\small
\centering
\begin{tabular}{lcccc}
\toprule
\multirow{2}{*}{Model}    & \multicolumn{2}{c}{Augmentation} & \multicolumn{2}{c}{w/ Refinement} \\ 
                                   \cmidrule(lr){2-3}                 \cmidrule(lr){4-5}
                                   & C10            & C100            & C10            & C100            \\ 
\midrule
\textit{PreActResNet-18} & & & & \\
Standard       & \bf94.63  & 75.79 & 94.48   & \bf76.06 \\
Mixup          & 95.92 & 78.95  & \bf96.26 & \bf80.85\\
Manifold Mixup & 95.81 & 80.19  & \bf96.07 & \bf81.39 \\
Cutmix         & 96.21  & 79.75  & \bf96.38 & \bf80.46 \\
AutoAugment    & 96.02  & 79.33 & \bf96.20 & \bf80.09 \\
\midrule 
\multicolumn{2}{l}{\textit{WideResNet-28-10}} & & & \\
Standard       & 96.11 & 81.15 & -  & - \\
Mixup          & 97.05 & 82.11  & \bf97.52 & \bf84.25 \\
Manifold Mixup & 97.13 & 83.02  & \bf97.32 & \bf84.77 \\
Cutmix         & 97.11 & 82.75  & \bf97.24 & \bf83.40 \\
AutoAugment    & 97.59 & 83.85  & \bf97.73 & \bf85.24 \\
\midrule
\multicolumn{2}{l}{\textit{Shake-Shake (26 2x96d)}} & & &\\
Standard       & 97.00 & 82.87 & - & - \\
Mixup          & 97.80 & 84.22 & \bf98.02 & \bf85.19 \\
Cutmix         & 97.77 & 84.51 & 97.76 & \bf84.64 \\
AutoAugment    & 98.02 & 85.62 & 98.05 & \bf86.13 \\
\bottomrule
\end{tabular}
% \vspace{-0.3cm}
\caption{Classification accuracy ($\%$) on CIFAR10 and CIFAR100.}
\label{tab_cifar}
\end{table}

\vspace{0.2cm}
\noindent$\bullet$\quad\textbf{Quantitative Results}

In Table~\ref{tab_cifar}, the mean values are calculated in three independent experiments by the median of the last 10 epochs in each experiment for PreActResNet18 and WideResNet-28-10. Our methods show a consistent improvement with different data augmentation methods on various backbones. Especially for Mixup, refining the models with standard data augmentation improves the accuracy significantly on CIFAR100. While the networks searched by P-DARTS~\cite{chen2019progressive} achieves 97.81\% test accuracy after being trained with AutoAugment for 600 epochs on CIFAR10, our method achieves a \textbf{97.97\%} accuracy with 550 epochs using AutoAugment and 50 epochs for refinement. 

Specially, we also conduct the experiments that refine the models trained with the standard data augmentation without any data augmentation methods for PreActResNet-18. There is a 0.27\% accuracy gain on CIFAR100, while a 0.15\% drop on CIFAR10. On the other hand, for other intensive augmentation methods, removing all data augmentations during refinement shows no significant difference with the experiments that preserve the standard data augmentations on CIFAR100. Therefore, we suggest to preserve the standard data augmentation during refinement.

\vspace{0.2cm}
\noindent$\bullet$\quad\textbf{Qualitative Analysis}

In Table~\ref{tab_loss}, cross-entropy (CE) losses on clean and augmented data are calculated to quantify the distribution gap between clean and augmented data to some extent. During refinement, the $\hat{R}_\mathrm{aug}$ is calculated with standard augmented data, which reflects a small distribution gap with clean data. Mixup brings the most significant difference between clean and augmented data, which can explain the significant improvement with refinement for Mixup. Interestingly, AutoAugment achieves a low CE loss on cleaning training data, yet refinement still works well to achieve a lower CE loss on clean data. These results suggest that data augmentation indeed helps the model converge to a better region, which can be hard to be arrived for directly training on clean data, and the gap between clean and augmented data obstructs the further convergence.  

\begin{table}[!t]
\small
\centering
\begin{tabular}{lcccc}
\toprule
\multirow{2}{*}{Method}  & \multicolumn{2}{c}{Augmentation} & \multicolumn{2}{c}{w/ Refinement} \\
               \cmidrule(lr){2-3}  \cmidrule(lr){4-5}
               & $\hat{R}_\mathrm{aug}$    & $\hat{R}$  & $\hat{R}_\mathrm{aug}$   & $\hat{R}$  \\
\midrule
Standard       & 3.3  & 1.1 & -   & \bf1.0 \\
Mixup          & 1356 & 98  & 4.7 & \bf2.4 \\
Manifold Mixup & 1253 & 67  & 4.2 & \bf2.2 \\
Cutmix         & 785  & 14  & 1.9 & \bf0.8 \\
AutoAugment    & 245  & 0.9 & 1.5 & \bf0.8 \\
\bottomrule
\end{tabular}
% \vspace{-0.3cm}
\caption{Cross-entropy losses ($\times 10^{-3}$) of augmented and clean training data on CIFAR100 for \textit{PreActResNet-18}. $\hat{R}_\mathrm{aug}$ and $\hat{R}$ repectly refer to $\hat{R}(f|\mathcal{P}_\mathrm{aug})$ and $\hat{R}(f|\mathcal{P})$.}
\label{tab_loss}
\end{table}

\begin{figure}[h]
\centering
    {\includegraphics[width=8cm,height=2.7cm]{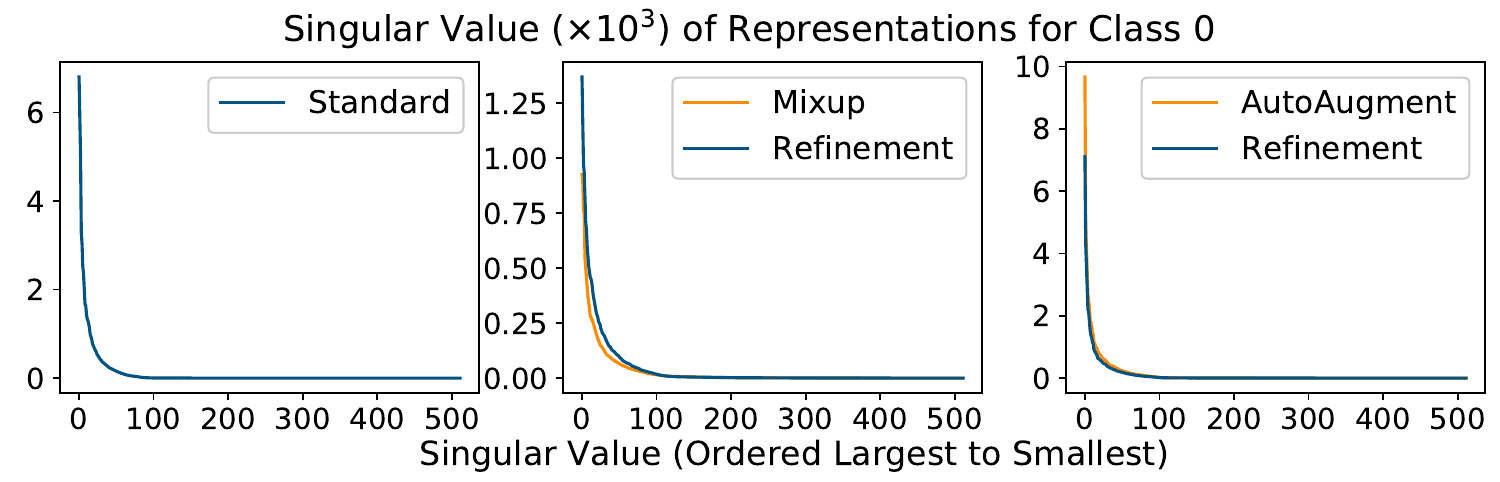}}\\
    {\includegraphics[width=8cm,height=2.7cm]{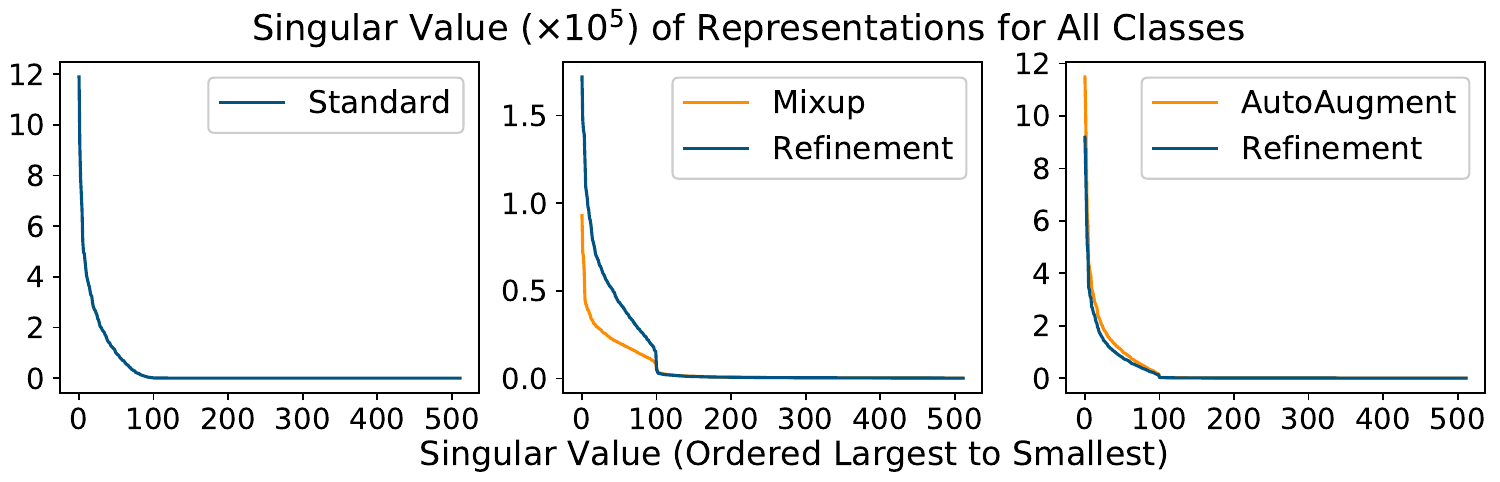}}
% \vspace{-0.3cm}
\caption{SVD on the representations in the penultimate layer of PreActResNet18 (512 neurons) on CIFAR100.}
\label{fig:singular}
\end{figure}

Since certain directions in the deep feature space imply meaningful semantic information~\cite{verma2018manifold,wang2019implicit}, we compute a singular value decomposition for the intra-class covariance matrix and the covariance matrix of all classes in the penultimate layer of PreActResNet-18 on CIFAR100 to analyze the variation of representations learned with different augmentation methods and refinement. The singular values are plotted and ordered form largest to smallest in Figure~\ref{fig:singular}. Here we only present the results of Mixup, since Manifold Mixup and Cutmix perform very similarly to Mixup.

We find that refinement tends to draw the distribution of singular values back to the original model to some extent for the intra-class covariance matrix. The singular values of models trained with Mixup-based methods and AutoAugment respectively are smaller and larger than the original model, while the singular values become closer to original values after refinement. It is interesting to find the different effects brought by Mixup-based methods and AutoAugment, which have not been noticed and researched as so far. A possible interpretation is that Mixup-based methods mainly affect the inter-class distances while AutoAugment introduces more features and induces invariance. 

Especially, for the covariance matrix of all classes, the curves for intensive data augmentation become steep near the 100th largest singular value. Notably, CIFAR100 has 100 categories and the last layer is a linear softmax classifier. The representation space will be more discriminative if the rank is larger than 100. However, compared with those larger singular values, the 100th largest singular value is almost 0 for the original model, which can be difficult for the classifier to learn. Such relative enlargement of singular values near the 100th largest can explain the effectiveness of these intensive data augmentation methods. After refinement, this characteristic is still preserved while the distribution of singular values is drawn back to some extent.

\vspace{0.2cm}
\noindent$\bullet$\quad\textbf{Ablation Studies}

\begin{figure}[!tb]
    \centering
    \includegraphics[width=8.2cm]{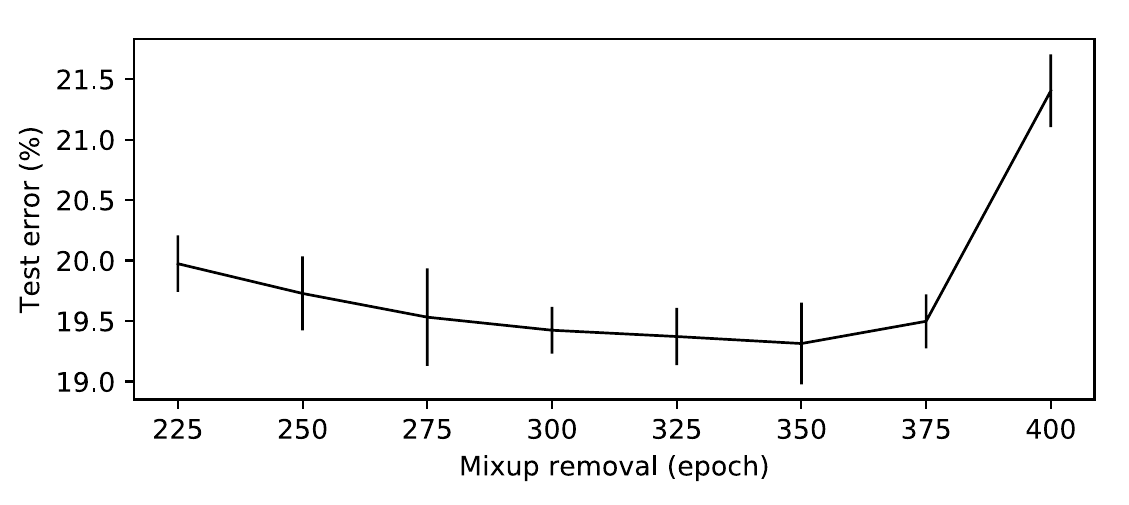}
    \caption{Test error (averaged over 5 runs) of PreActResNet-18 trained on CIFAR100 when Mixup is removed at different epochs (total of 400 training epochs). The bars represent the range of test errors for each number.}
    \label{fig_epoch}
\end{figure}

In previous experiments, we train models for a few more epochs to refine. Here we keep the total training epochs constant to verify the effectiveness of our method. We train PreActResNet-18 on CIFAR100 with Mixup, and the learning rate is divided by 10 at epochs 150 and 275. Figure~\ref{fig_epoch} shows the test error curve with different epochs, at which Mixup is removed, when a total of 400 training epochs. Especially, Mixup removal at 400 epoch means no refinement is performed. Besides, if we train models on CIFAR with intensive data augmentation for a fixed number of epochs, refining epochs will not influence results once convergence. 

% According to Figure~\ref{fig_epoch}, it seems that refinement would benefit more from the longer training epochs with intensive data augmentation. 
We also find that increasing the number of training epochs with intensive data augmentation benefits the performances after refinement, even though AutoAugment causes an obvious over-fitting on augmented data with long training epochs. Table~\ref{tab_auto} shows refinement improves accuracy significantly, and suppresses the over-fitting on augmented data. For PreActResNet-18 of $T_1=1000$, the learning rates divided by 10 at epochs 400 and 800. For WideResNet-28-10, Cosine learning rate decay is implemented. 

\begin{table}[!tb]
\small
\centering
\begin{tabular}{lcc}
\toprule
Model         & AutoAugment & w/ Refinement \\
\midrule
\textit{PreActResNet-18} & & \\
$T_1=400$, $T_2=50$    & 79.33         & \bf80.09      \\  
$T_1=1000$, $T_2=200$  & 78.16         & \bf80.38      \\ 
\midrule
\textit{WideResNet28-10} & & \\
$T_1=200$, $T_2=50$    & 83.73         & \bf84.06       \\ 
$T_1=400$, $T_2=50$    & 83.85         & \bf85.24       \\ 
\bottomrule
\end{tabular}
% \vspace{-0.3cm}
\caption{Classification accuracy ($\%$) for different training epochs with AutoAugment on CIFAR100. $T_1$ and $T_2$ refer to the number of epochs with AutoAugment and the number of refinement epochs, respectively.}
\label{tab_auto}
\end{table}

Moreover, we try to weaken the intensity of data augmentation gradually so that refining models from augmented data to clean data gradually. For Mixup, $\gamma$ is decreased to 0 gradually. For Cutmix and AutoAugment, we decrease the probability to implement data augmentation gradually. The results show no significant difference with the earlier experiments. 

\subsection{Results on Tiny-ImageNet}
Tiny-ImageNet consists of 200 $64\times64$ image classes with 500 training and 50 validation per class. We train PreActResNet-18 for 400 epochs with intensive data augmentation, and refining the models for 50 epochs. Other hyper-parameters about model training are the same as the settings in previous experiments.

The augmentation policies found by AutoAugment is searched on CIFAR10 and ImageNet. Here we implement both CIFAR-policy and ImageNet-policy on Tiny-ImageNet. Following the setting in the paper~\cite{Cubuk_2019_CVPR}, we apply Cutout with $32\times32$ pixels after CIFAR-policy. The results are listed in Table~\ref{tab_tiny}, including in different distributions $\lambda \sim \mathrm{Beta}(\gamma, \gamma)$ for Mixup and Manifold Mixup and different probabilities $p$ to implement Cutmix. Consistent gains are achieved over various data augmentation methods. It is worth emphasizing that greater improvements on the last results than the best results in most cases, which means refinement alleviates the over-fitting on augmented data.

\begin{table}[t]
\small
\centering
\begin{tabular}{lcccc}
\toprule
\multirow{2}{*}{\textit{PreActResNet-18}}    & \multicolumn{2}{c}{Augmentation} & \multicolumn{2}{c}{w/ Refinement} \\ 
                                   \cmidrule(lr){2-3}                 \cmidrule(lr){4-5}
                                   & Best            & Last            & Best            & Last            \\ 
\midrule
Standard                           & 60.84           & 60.68           & -           & -           \\ 
\midrule
Mixup & & & & \\
$\gamma=0.2$                     & 63.18           & 63.08           & \bf64.54           & \bf64.20           \\ 
$\gamma=0.5$                     & 63.95           & 63.34           & \bf65.45           &     \bf65.08       \\ 
\midrule
Manifold Mixup & & & & \\
$\gamma=0.2$                     & 63.66           & 63.28           & \bf64.54           & \bf64.33           \\ 
$\gamma=0.5$                     & 64.88           & 64.43           & \bf65.98           &     \bf65.80       \\
\midrule
Cutmix & & & & \\
$p=0.5$                          & 64.90           & 64.61           & \bf65.84           &    \bf65.59        \\
$p=1  $                          & 65.97           & 65.23           & \bf66.29           & \bf65.87           \\
\midrule
AutoAugment & & & & \\
CIFAR-Policy                     & 65.08           & 64.29           & \bf65.31           & \bf65.12           \\
ImageNet-Policy                  & 61.06           & 60.65           & \bf61.82           & \bf61.75           \\
\bottomrule
\end{tabular}
% \vspace{-0.3cm}
\caption{Classification accuracy ($\%$) on the validation set of Tiny-ImageNet. Both best and last results are reported.}
\label{tab_tiny}
\end{table}

\subsection{Results on ImageNet (ILSVRC2012)}\label{sec_imagenet}
On the ILSVRC2012 classification dataset~\cite{russakovsky2015imagenet}, we train models with initial learning rate 0.1 and a mini-batch size of 256 on 8 GPUs and follow the standard data augmentation: scale and aspect ratio distortions, random crops, and horizontal flips. For Mixup, the models are trained for 200 epochs, and the learning rate is divided by 10 at epochs 60, 120, 180. For AutoAugment, ResNet-50 is trained for 300 epochs, and the learning rate is divided by 10 at epochs 75, 150, and 225, while ResNet-101 is trained for 200 epochs. We refine all models with standard data augmentation for 20 epochs by a learning rate of $10^{-4}$. In Table~\ref{tab_image}, ResNet-50 for Mixup of $\gamma=0.5$ performs worse than Mixup of $\gamma=0.2$, however, they achieve similar accuracy after being refined. Since $\gamma$ is to control the strength of interpolation, which can be understood to control the distribution gap, such results reflect the refinement also helps to weaken the negative impacts of the distribution gap.

\subsection{Transferring to Object Detection}
To verify the effectiveness of our approach further, we conduct experiments on object detection in the PascalVOC 2007 dataset~\cite{everingham2010pascal} by the pre-trained ResNet-50 that are trained in section~\ref{sec_imagenet}. The all detection models are trained following the standard strategy with Faster R-CNN~\cite{ren2015faster} whose backbone is initialized with the pre-trained ResNet-50. The models are trained for 12 epochs with a mini-batch size of 16, and the learning rate starts at 0.01 and is divided by 10 after 9 epochs. 

In Table~\ref{tab_detect}, the pre-trained backbone models of Mixup and AutoAugment fail to improve the performance on object detection task over the original model, although they perform well on the classification task in ILSVRC2012. Interestingly, the pre-trained models with refinement achieve higher mAP than not only the models without refinement but also the original model. Here, we try to give a reasonable explanation for this. Different from classification, detection needs additive features about location. However, these intensive augmentation designed only for classification could treat the features about location as minor features and ignore them, as mentioned in section~\ref{sec_converge}. After refinement, the features about location are learned more precisely, which can help models transfer to detection. This means the complicated data augmentation methods, which were applied in classification and viewed as inappropriate for detection before, can also benefit detection models with our approach.

\begin{table}[!tb]
\small
\centering
\begin{tabular}{lcccc}
\toprule
\multirow{2}{*}{Model}     & \multicolumn{2}{c}{Augmentation} & \multicolumn{2}{c}{w/ Refinement} \\ 
          \cmidrule(lr){2-3}                 \cmidrule(lr){4-5}
                           & Top-1           & Top-5          & Top-1           & Top-5    \\
\midrule
\textit{ResNet-50} & & & & \\
Standard                           & 76.39           & 93.19           & -               & -               \\ 
Mixup ($\gamma=0.2$)               & 77.47           & 93.75           & \bf77.69           & \bf93.83           \\ 
Mixup ($\gamma=0.5$)               & 77.26           & 93.78           & \bf77.65           & \bf93.93           \\ 
%Cutmix ($p=1$)                     & 77.94           & 93.94           & 77.82           & 93.89           \\
AutoAugment                        & 77.83           & 93.70           & \bf77.98           & \bf93.86           \\
\midrule
\textit{ResNet-101} & & & &\\
Standard                           & 78.13           & 93.71           & -               & -               \\ 
Mixup ($\gamma=0.5$)               & 79.41           & 94.70           & \bf79.61           & \bf94.73           \\ 
AutoAugment                        & 79.20           & 94.45           & \bf79.33            & \bf94.46          \\
\bottomrule
\end{tabular}
% \vspace{-0.3cm}
\caption{Classification accuracy ($\%$) on ImageNet.}
\label{tab_image}
\end{table}

\begin{table}[]
\small
\centering
\begin{tabular}{lcc}
\toprule
Method       & Augmentation     &w/ Refinement     \\
\midrule
Standard       & 72.0     & - \\
Mixup          &    67.4  & \bf73.3 \\
AutoAugment    &   70.5   & \bf73.1\\
\bottomrule
\end{tabular}
\caption{mAP ($\%$) on PascalVOC 2007 object detection, obtained by training the pre-trained \textit{ResNet-50} on ILSVRC2012 with Faster R-CNN. The augmentation and refinement refer to different training methods on ILSVRC2012.}
\label{tab_detect}
\end{table}

\section{Conclusions}

This paper presents a simple yet effective approach for network optimization, which adopts (usually complicated) augmentation for generating abundant training data, but switch off these intensive data augmentation to refine the model in the last training epochs. In this way, the model often arrives at a reduced testing loss, with the generalization error and empirical loss balanced. We also show intuitively that augmented training enables the model to traverse over a large range in the feature space, while refinement assists it to get close to a local minimum. Consequently, models trained in this manner achieve higher accuracy in a wide range of visual recognition tasks, including in a transfer scenario.

Our work sheds light on another direction of data augmentation which is complementary to the currently popular trend that keeps designing more complicated manners for data generation. It is also interesting to combine refined augmentation with other algorithms, \textit{e.g.}, a cosine-annealing schedule for refinement, or add this option to the large space explored in automated machine learning.

{\small
\bibliographystyle{ieee_fullname}
\bibliography{egbib}
}

\end{document}